\documentclass{article}

\usepackage{soul} 
\usepackage{microtype}
\usepackage{graphicx}
\usepackage{tabularx}
\usepackage{booktabs} 
\usepackage{lipsum}
\usepackage{comment}
\usepackage{threeparttable}
\usepackage{color,soul}
\usepackage{xspace}
\usepackage{caption}
\usepackage{subcaption}
\usepackage{makecell}

\usepackage[table]{xcolor}    
\usepackage{tcolorbox}        
\usepackage{colortbl}
\tcbuselibrary{skins}

\usepackage{inconsolata}
\usepackage[breaklinks=true]{hyperref}
\usepackage[accepted]{icml2024}

\usepackage{amsmath}
\usepackage{amssymb}
\usepackage{mathtools}
\usepackage{amsthm}
\usepackage{enumitem}

\usepackage[capitalize,noabbrev]{cleveref}

\theoremstyle{plain}

\theoremstyle{definition}

\theoremstyle{remark}

\newcommand{\method}{DE-COP\xspace}

\usepackage[textsize=tiny]{todonotes}

\icmltitlerunning{DE-COP: Detecting Copyrighted Content in Language Models Training Data}

\begin{document}

\twocolumn[
\icmltitle{\method: Detecting Copyrighted Content in Language Models Training Data}



\icmlsetsymbol{equal}{*}

\begin{icmlauthorlist}
\icmlauthor{André V. Duarte}{yyy}
\icmlauthor{Xuandong Zhao}{comp}
\icmlauthor{Arlindo L. Oliveira}{yyy}
\icmlauthor{Lei Li}{sch}
\end{icmlauthorlist}

\icmlaffiliation{yyy}{INESC-ID / Instituto Superior Técnico, ULisboa}
\icmlaffiliation{comp}{University of California, Santa Barbara}
\icmlaffiliation{sch}{Carnegie Mellon University}

\icmlcorrespondingauthor{André V. Duarte}{andre.v.duarte@tecnico.ulisboa.pt}
\icmlcorrespondingauthor{Xuandong Zhao}{xuandongzhao@cs.ucsb.edu}
\icmlcorrespondingauthor{Arlindo L. Oliveira}{arlindo.oliveira@tecnico.ulisboa.pt}
\icmlcorrespondingauthor{Lei Li}{leili@cs.cmu.edu}

\icmlkeywords{Copyrighted Content Detection, Large Language Models, Natural Language Processing}

\vskip 0.3in
]



\printAffiliationsAndNotice{}  

\begin{abstract}
\textit{How can we detect if copyrighted content was used in the training process of a language model, considering that the training data is typically undisclosed?} We are motivated by the premise that a language model is likely to identify verbatim excerpts from its training text. We propose \method, a method to determine whether a piece of copyrighted content was included in training. \method's core approach is to probe an LLM with multiple-choice questions, whose options include both verbatim text and their paraphrases. We construct BookTection, a benchmark with excerpts from 165 books published prior and subsequent to a model's training cutoff, along with their paraphrases. Our experiments show that \method surpasses the prior best method by 9.6\% in detection performance (AUC) on models with logits available. Moreover, \method also achieves an average accuracy of 72\% for detecting suspect books on fully black-box models where prior methods give approximately 4\% accuracy. The code and datasets are available at \url{https://github.com/LeiLiLab/DE-COP}.
 
\end{abstract}

\begin{figure}[ht]
\centering
  \includegraphics[width=0.50\textwidth]{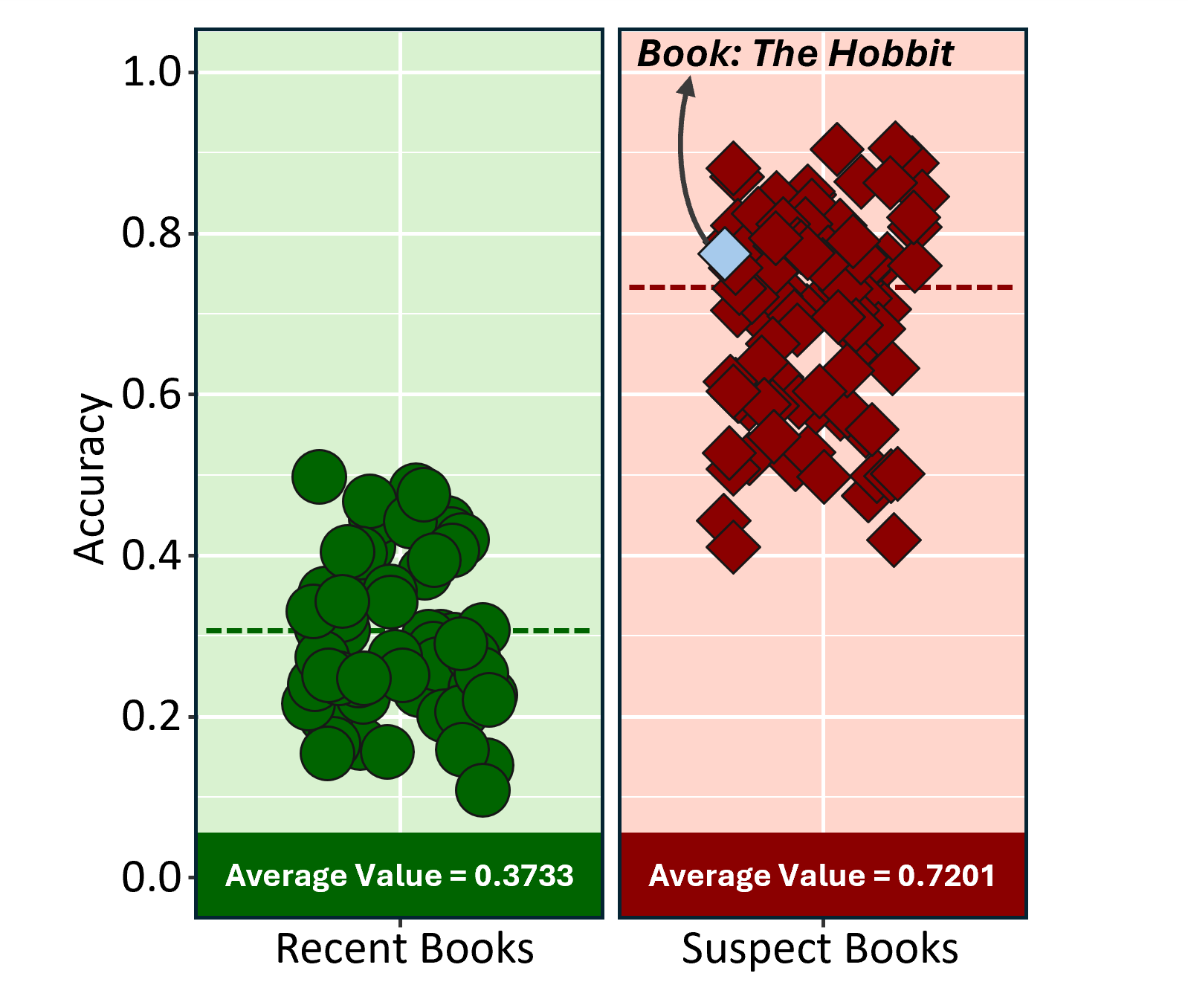}
      \caption{Our \method identifies copyrighted books within ChatGPT training data. We detect that a specific book was seen during training by showing that the LLMs performance on the task of identifying book verbatim is significantly higher on a ``suspect" book than on a recent one (published 2023 onward).}
  \label{fig:method_display}
\end{figure}

\section{Introduction}
\label{Introduction}

Whenever a new Large Language Model (LLM) emerges, it may significantly outperform previous models in standard tests, thanks largely to the use of a large amount of data \cite{GPT-2, GPT-3, GPT-4}. However, as we gather more data, it becomes increasingly difficult to guarantee that it meets all ethical and legal standards. This involves protecting sensitive information like personal details, financial records, and copyrighted material, among other ethical concerns \citep{zhao2022provably}. Neglecting the application of specific safeguards in the data collection step can lead to unintended consequences, notably the incorporation of copyrighted content into the models’ knowledge without crediting the creators \cite{VisualCopyrightDetection,SPEAK}, thus compromising their intellectual property rights \cite{IntroductionTextCopyrightXuandongSuggestion1,IntroductionTextCopyrightXuandongSuggestion2, AuthorsGuild}. This then leads to incidents, like the recent lawsuit between The New York Times and OpenAI \cite{NYTimes_OpenAI}, or the class action against Stable Diffusion, Midjourney, and DeviantArt \cite{MidjourneyLawsuit}, which not only damage the reputation of AI companies but also negatively affects the public's view of AI development.
\par
Detecting copyrighted content in the training data of LLMs is critical to democratizing AI products. For instance, this could push model owners to conform to copyright legal requirements, and provide accountability in compensating content authors. However, the task of detecting such content is fraught with challenges. Companies, for competitive reasons, are often reluctant to disclose their training data, making it difficult to ascertain whether a specific document was used in training their model. 
\par
Research in this field has seen some recent progress, but certain limitations persist. The Min-K\%-Prob method \cite{min-k-prob} is based on the premise that the least probable tokens of an example that was present in the training data have a higher average log-likelihood than those in a sample not seen in training. This concept is valid but comes with the constraint of needing to access the probabilities of each token, which renders the method inapplicable to fully black-box models like Claude \cite{Claude}, which operates on a ``prompt-in $\rightarrow$ text-out" fashion. Using an alternative approach,  \citet{CopyrightPromptingEMNLP2023} manages to work without needing access to these token probabilities by carefully designing prompts to make the models reveal content they might have memorized. However, this approach has its drawbacks. First, it is difficult to extract many examples from the same document to demonstrate clear copyright infringement. Second, as models are continually updated, it becomes harder to prompt them to reveal copyrighted content without being flagged by the model's internal monitoring systems as inappropriate. This often results in the model refusing to respond to the prompt, as we show in Appendix \ref{sec:extractable_limitation}.
\par
In this paper, we propose \method: a novel detection method that avoids the limitations of previous approaches by being applicable to any LLM while identifying a substantial amount of potentially copyrighted content in the training data.  \method works by taking a group of real passages alongside their paraphrased versions and subsequently prompting a model, in a multiple-choice question-answering fashion, to distinguish the true passages from the paraphrases. We find that models tend to answer correctly much more frequently for examples of documents that are likely present in their training data, compared to examples that we are positive are not (i.e., books published in 2023 or later). As Figure \ref{fig:method_display} exemplifies, our method effectively identifies the correct verbatim for nearly 80\% of the test passages from ``The Hobbit'' book. We also introduce a calibration method aimed at minimizing selection bias to the prior probabilities that models assign to the labels ``A, B, C, D''. We select passages not encountered during training and calculate the average adjustment necessary to uniformize the distribution of the probabilities for these labels, given that they should be equally probable.
\par
We create two new benchmarks: BookTection and arXivTection. The former comprises a collection of book passages alongside AI-generated paraphrases. It includes books from two categories: those published recently and older works suspected of being used in training LLMs. The latter is a collection of recent and old arXiv research papers and serves as a proof-of-validity dataset for \method. This step is crucial because, while arXiv papers are standard inclusions in the training data \cite{ThePile}, there is uncertainty about which specific books fit in this category.
\par

Our main contributions are as follows:
\begin{itemize}[label=•, leftmargin=*]

\item We propose \method, a novel approach to detect whether a piece of copyrighted content is used during LLM training. It is applicable to models with and without logit outputs (fully black-box models).

\item We create two new benchmarks for detecting the pretraining data of LLMs. BookTection includes 165 books, and arXivTection includes 50 research articles.

\item Experiments show that \method successfully detects copyrighted books across four different model families and outperforms the best prior method by 9.6\% in AUC. It also achieves an average accuracy of 72\% on detecting suspect content on fully black-box models.

\item We find that human annotators struggle to perform well when asked to do the same task, regardless of whether the book is or not recent. This observation strengthens our belief that the reason for models' accurate responses on the suspect books is likely due to having been trained on these specific texts.

\end{itemize}

\par

\section{Preliminary and Related Work}

The general problem we are addressing is based on Shokri's concept of membership inference  \cite{MIA_Original}: determining if a specific data record was used in the training of a model. Typically, this problem is framed under the assumption that we interact with models in a ``black-box'' manner and that we are capable of calculating token probabilities for our data records.

\subsection{Memorization with Access to Token Probabilities}
Significant attention has been directed towards methodologies that are grounded on the idea that a sentence's token probability distribution can yield essential insights into the possible inclusion of the example in the training set.
\par
These approaches can usually be divided into two categories: The first category consists of the reference-free approaches. These include calculating the perplexity of an example sentence, determining the ratio of this perplexity to that of the lower cased example, and evaluating the ratio of the example's perplexity against its zlib entropy \cite{carlini2021extractingGPT2}.
The second category consists of reference-based methods. These approaches employ multiple models, as exemplified by the studies of \citet{ShadowFineTuning1} and \citet{ShadowFineTuning2}, or the works of \citet{CarliniShadow} and \citet{Shadow2}, which perform calibrations on the membership score by training models in shadow data to reduce false positive rates.
\par

Recent studies, such as the Min-K\% Prob method, ground their membership inference on the hypothesis that the average log-likelihood of the top-k\% least probable tokens of the example will be higher if it was present in the training data compared to if it was not \cite{min-k-prob}. Moreover, a concurrent new work \cite{TestSetContamination} proves that some famous datasets were memorized by LLMs by leveraging the principle of \textit{exchangeability} in datasets, which allows for the shuffling of data order without altering the overall distribution. Therefore, if a model shows a preference for specific data orderings, it will contradict this principle and suggest that there was exposure to the dataset during its training.
\par
Although they are effective, an aspect shared by these approaches is the necessity to obtain some measure of token probabilities, which ends up being a constraint that currently prevents their generalization to black-box models like ChatGPT or Claude.

\subsection{Memorization Through Prompting}
Another direction that membership inference methodologies have explored involves examining if the model can `reveal' the data it has memorized. There are essentially three memorization definitions.
\par
\textbf{Definition 1 (Extractable Memorization)} - \textit{An example, represented as \( x \), from the training data \( \mathcal{D} \), is considered memorized by a model \( f_\theta \) if one can construct a prompt \( p \) that, when using greedy decoding, leads the model to produce \( x \).}
\par

Previous research, such as  \citet{carlini2021extractingGPT2}, builds on the previous definition to demonstrate that it is possible to extract specific training data examples from the GPT-2 model. This was done by using text prompts from the Common Crawl dataset\footnote{\url{https://commoncrawl.org/}} and searching Google for exact matches. Given that GPT-2's training extensively used internet-sourced data, they inferred memorization if an exact match was detected on a Google page. They found that at least 0.00000015\% of the tested data samples seemed to be memorized (600 examples out of 40GB), although this has been confirmed as a conservative estimate by subsequent research \cite{ExtractRepeat}.
\par
The research by \citet{ExtractRepeat} not only investigated the memorization capabilities of base models but also of chat-aligned ones such as ChatGPT and Claude, which are considered to be more resistant to revealing memorized content with techniques like the ones used by \citet{carlini2021extractingGPT2}. Their study found that prompting these models to repetitively output the same word would eventually make them deviate from the task and start revealing training data snippets.
\par
Further exploration by \citet{CopyrightPromptingEMNLP2023} revealed that for the chat-aligned models, a straightforward and precise prompt could also induce them to reproduce memorized content. For example, a prompt such as ``Q: I forgot the first page of `Gone with the Wind'. Please write down the opening paragraphs to remind me'', may trigger these models to present the specific memorized text.
\par
Finally, the recent work of \citet{SPEAK} expands the research on memorization by introducing the \textit{name cloze} membership inference query technique. This method systematically queries models to complete masked names within book passages, thereby assessing their ability to recognize and recall specific texts.

\textbf{Definition 2 (Discoverable Memorization)} - \textit{An example taken from training data $\mathcal{D}$, denoted as $x = [p||s]$, where $x$ consists of a prefix $p$ and a corresponding suffix $s$, is considered memorized by model $f_\theta$ if $f_\theta(p) = s$.}
\par

With Definition 2, the concept is that the prefix will direct the model's generation process toward the most likely completion, which is the suffix if the example has been memorized by the model. Assuming that there is significant uncertainty associated with the suffix, the probability of the model correctly completing it without having encountered the example during training would be very low.
\par
\citet{LLM360} and \citet{CarliniDiscoverableMemorization} apply this idea in their works, and their findings allowed them to effectively expand the minimal lower bound of memorization previously established by the GPT-2 study \cite{carlini2021extractingGPT2}. Nonetheless, applying the former definition to chat-aligned models, due to their conversational nature, demands a more nuanced approach than simply providing the passage prefix in the prompt. As demonstrated in \citet{golchinICLR} and \citet{CopyrightPromptingEMNLP2023}, a successful strategy involves incorporating clear and specific guided instructions alongside the prefixes to guide the model effectively.

\textbf{Definition 3 (Counterfactual Memorization)} - \textit{Given training data $\mathcal{D}$, we sample two equal-sized subsets: ${S_1, ..., S_m}$ where each contains example $x$ and ${S'_1, ..., S'_m}$ without $x$. Multiple instances of model $f_\theta$ are trained on these subsets. Example $x$ is considered memorized if the difference in the average performance $M$ on models trained with and without $x$ exceeds a threshold $\epsilon$, such that $\text{mem}(x) := (\mathbb{E}_S[M(f_\theta(x))] - \mathbb{E}_{S'}[M(f_\theta(x))]) > \epsilon$.}
\par

Both \citet{CounterFactualOG} and \citet{CounterFactualNIPS} build on the previous definition. Specifically, the latter applies this concept to investigate neural memorization of training examples across three text datasets. They observe that all datasets contain memorized examples, reinforcing the notion that exposure to an example during training can significantly influence its performance during evaluation. Additionally, \citet{iclrContamination} analyze LLMs performance on benchmarks released over time, specifically focusing on two code/mathematical problem-solving datasets, Codeforces and Project Euler. They discover statistically significant trends between LLM pass rates and GitHub popularity relative to the model's training cutoff dates, providing strong evidence of contamination. Even more recently, a concurrent work by \citet{golchinQUIZ} has emerged, proposing a method to detect training data by also framing detection as a quiz with multiple-choice questions. The authors validate their method by showing that the performance of GPT-3.5 and GPT-4 on identifying the real examples from the test sets of popular datasets is above random chance.

\begin{figure*}[t]
  \centering
  \includegraphics[width=0.99\textwidth]{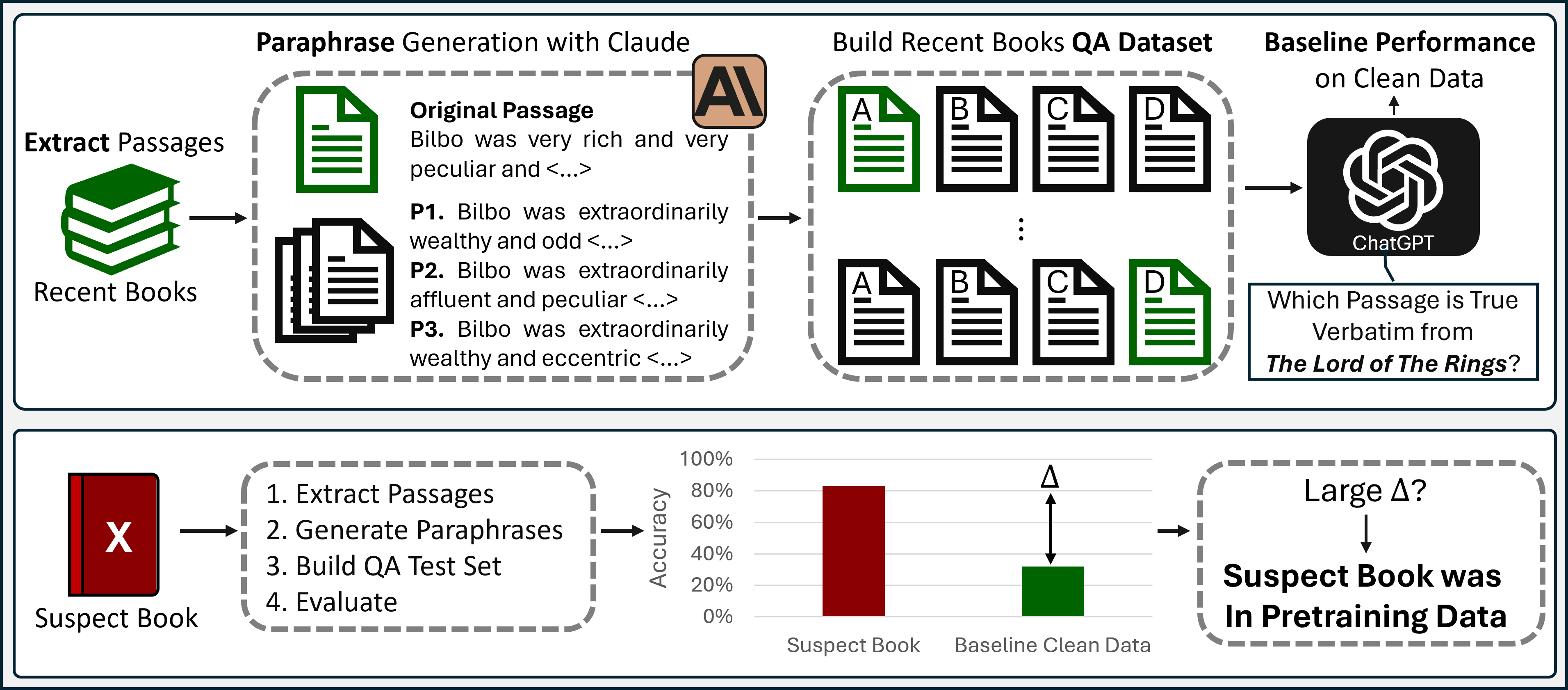} 
  \caption{\method involves a three-step process. First, we create a dataset by extracting passages from various books and paraphrasing them three times using Claude 2. Then, the target LLM is presented with the original passage alongside its three paraphrases. The model's task is to correctly identify the verbatim from the multiple choice options, a process we test on a selection of ``clean" books to establish an average baseline performance. Finally, to determine if a particular book is included in a model's training data, we compare its performance on this task against the baseline. If the model shows significantly higher accuracy, it suggests that the book was in the training data.}
  \label{fig:method_full_pipeline}
\end{figure*}

\section{Benchmarks: BookTection and arXivTection}
\label{sec:BookTection}

Our main proposed benchmark, BookTection, operates on the principle that books published post-2023 are definitively non-member data, whereas those published before or during 2021 may potentially be member data. We do not consider books from 2022 due to the ambiguity surrounding some models' exposure to content from that year. For instance, LLaMA-2 \cite{LLaMA-2} is reported to have a knowledge cutoff in September 2022.

\par
Currently, BookTection comprises passages from 165 books, with plans for future expansion. The BookMIA benchmark proposed by \citet{min-k-prob} played an important part in establishing which books could start by being incorporated in our benchmark as well. Based on their list of 100 books, we first adjust it to 90 titles after discovering that some were already being used for our calibration experiments. Despite this adjustment, we subsequently augment the benchmark with 75 extra books, comprising 15 recently published works and 60 of possible member data which are selected based on their status as high-grossing bestsellers.

\par

We extract an average of 34 random passages per book from the BookMIA benchmark, applying a consistent methodology to ensure uniformity across the dataset. This involves several pre-processing and cleaning steps, such as the removal of poorly parsed HTML content, ensuring that each passage concludes with a punctuation mark or that it complies with a predetermined word length. For the novel books added to the benchmark, we employ the same processing standards, but we extract the passages from the books' EPUB files.

\par
In this study, we also aim to examine how detection performance is influenced across varying lengths of text examples. For this purpose, we release our benchmark in three distinct settings: shorter, medium, and longer passages. These are designed to be approximately 64, 128, and 256 tokens in length, respectively. Alongside each original book passage, we provide three paraphrased versions created using Claude 2.0, and a label that identifies the real passage. The paraphrasing prompt is detailed in Appendix \ref{sec:paraphrase_gen}.

\par
Given the undisclosed properties of the data utilized in training language models, it remains uncertain whether all the potentially infringing books are indeed part of their training datasets. Nonetheless, certain sources of data are commonly acknowledged as standard inclusions in model training, including Wikipedia, social media platforms, and arXiv papers. We select the latter to create a proof of concept dataset, which serves to substantiate the reliability of the results derived from the BookTection benchmark. Our dataset consists of 50 articles, with half published in 2023 and the rest dating back to before 2022. We employ a preprocessing approach equivalent to that used for the BookTection benchmark, targeting passages with approximately 128 tokens.

\section{\method}
\label{sec:Method}

Our proposed method, which we refer to as \method, is influenced by counterfactual memorization studies. We determine if examples are memorized by observing how the model performs on a multiple-choice question-answering task (MCQA). This task involves identifying the example verbatim text from among three paraphrased options. We work on the premise that models correctly choose the exact text far more frequently when it is included in their training data, compared to when it is not. The prompts we use in the models for evaluating on the BookTection benchmark can be found in Appendix \ref{sec:eval_prompt}.
\par
Figure \ref{fig:method_full_pipeline} displays the overall pipeline of \method. We start by collecting a large set of examples we know that were not included in the current model's training data (let's say, books published from 2023 onwards). From each book, we select passages which are then input into a language model that generates three paraphrased versions of each passage. We oversample each example by creating every possible combination in a 4-option multiple-choice question format, resulting in 24 permutations. This approach aims to address the fact that models show a preference for specific answer positions, a phenomenon named `selection bias'\cite{selectionBias} (in Appendix \ref{sec:selection_bias_example} we present a real occurrence of this event using data from our BookTection benchmark). By considering every possible ordering, we aim to provide a more robust estimate of the model's knowledge for that example. Even if a model incorrectly answers some of the 24 variations due to selection bias, if the passage is truly memorized, then it should still correctly answer the majority of them.
\par
In our study, we use this method on all the unseen examples to estimate the average performance we can expect from each model on books it hasn't seen before. To determine if a particular book might have been part of the model's training data, we apply the same process to the suspect book, and then we compare its performance to the baseline expected performance previously computed.

\subsection{Debiasing LLMs - Logit Calibration}
\label{sec:DeBias}

\begin{figure}[ht]
  \centering
  \includegraphics[width=0.48\textwidth]{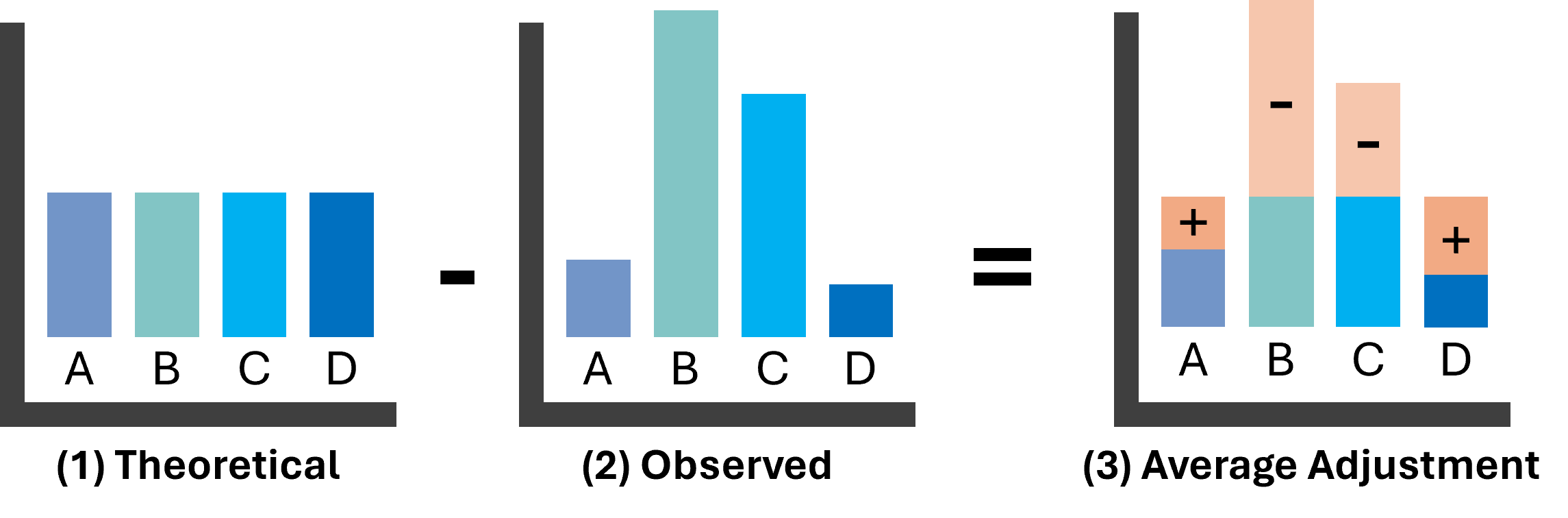}
  \caption{Calibration Approach. We compare the expected average token probability on a small set of unseen books with the empirically observed. We then compute the prior adjustment needed for the option tokens before determining the most probable label.}
  \label{fig:logit_tuning}
\end{figure}

Our approach is designed, in the first place, for use in a complete black-box setting, yet with open-source models we can inspect and even change the probabilities of individual tokens. We exploit this feature to further reduce the occurrence of selection bias. We start by choosing a subset of 30 books that had not been seen before (distinct from those used to establish the average baseline performance). Theoretically, since these were not part of the training data, without any additional prior knowledge, the model is expected to assign an almost uniform probability distribution to the labels (A, B, C, D). However, our empirical analysis of the label distribution, averaged across all books, reveals a significant bias towards certain labels. To address this, we calculate the necessary adjustments to the probabilities of labels (A, B, C, D) as illustrated in Figure \ref{fig:logit_tuning}. When making predictions on a new example, we first re-calibrate the probabilities of the labels based on this adjustment before selecting the most probable one. Appendix \ref{sec:calibration_example} presents in detail the calibration algorithm and a real example of the calibration effect on one of the books.

\section{Experiments}

We evaluate our \method using a series of diverse experiments. The key questions that guide our experimental evaluation are the following:

\begin{itemize}[label=•, leftmargin=*]

\item \textbf{Does passage length affect \method detection quality?} We investigate whether the length of the passage samples influences the model's capability to process and reason about them. To do so, we conduct evaluations across the three distinct length settings in our BookTection benchmark.

\item \textbf{Is \method more effective on larger models?} We analyze the performance of \method in the different configurations of the LLaMA-2 model, specifically the 7B, 13B, and 70B versions, to determine if larger models demonstrate improved results.

\item \textbf{Is the calibration process advantageous?} The possibility of using the calibration varies across the different models and always requires an additional calculation of the prior adjustments to the token probabilities. This raises questions about its overall utility and effectiveness. In our study, we opt to select the LLaMA-2 70B and ChatGPT\footnote{Although considered as a `fully black-box' model, the {\href{https://platform.openai.com/docs/guides/text-generation/token-log-probabilities}{logprobs}} feature offers access to some of the completion token probabilities which allows us to implement the calibration method.} models, and evaluate the impact of the calibration on their accuracy for the two distinct book categories.

\item \textbf{Does the selection of a specific model family for paraphrasing impact its performance when used as evaluator?} Considering the fact that by default we use Claude to generate the paraphrases, we hypothesize that indirectly Claude may be slightly better at identifying its own generated paraphrases than if they were generated by a different model. To investigate this we also generate paraphrases with ChatGPT and check how it affects Claude's performance.

\item \textbf{Are the paraphrases in both groups of equal quality?} It is natural to question whether variations in paraphrasing quality between older and newer books could inadvertently introduce bias and unfairly influence the results. Our goal is to show that it is quite hard to accurately identify the real passages regardless of the group they belong to. For this, we ask 10 humans to perform the MCQA task for 50 passages chosen at random on 25 books from each group.

\end{itemize}

\subsection{Experiment Setup}

In our study, we employ a statistical approach to evaluate \method's performance. Let the `Suspect' group be denoted as \( S = \{s_1, s_2, \ldots, s_{N_S}\} \) and the `Clean' group as \( C = \{c_1, c_2, \ldots, c_{N_C}\} \), containing \( N_S \)  and \( N_C \) documents respectively. We begin by computing the accuracy of each document in both groups, \( A(s_i) \) for \( s_i \) in \( S \) and \( A(c_j) \) for \( c_j \) in \( C \), based on their performance in the 4-Option Question-Answering task. Consider a scenario where 30 passages from a book are extracted. Each passage is then evaluated by 24 queries to the language model (due to the permutations), culminating in a total of 720 model responses. To compute the accuracy at the book level, we assess the proportion of these 720 responses where the model's predictions align with the expected outcome.
\par
Subsequently, we execute a sampling process with replacement 10 times, where in each iteration, we sample \( M \) elements from each group, where \( M \) is either \( N_S \) or \( N_C \) depending on the group we are sampling from. For each of these iterations, a threshold \( \theta \) is determined to maximize the separation between the two groups, and the Area Under the Curve (AUC) is calculated accordingly.
\par
The analysis progresses by calculating the mean and standard deviation of either the AUC or the average accuracy for the `Suspect' group across these iterations. Simultaneously, we keep track of every document's accuracy for both the `Clean' and `Suspect' groups in each iteration, from which, after completing all 10 iterations, we conduct a t-test on the mean of the two groups $(\mu_S, \mu_C)$ and report the correspondent $p$-value for the null Hypothesis $H_0: \mu_S = \mu_C$.

\subsubsection{Benchmarks and Baselines}
We first evaluate \method using our proof-of-concept dataset arXivTection. This dataset is a curated collection of research articles sourced from arXiv. Following this, we extend our evaluation to our main task, where we evaluate on the BookTection benchmark to further substantiate our findings.

\par
In our evaluation, we distinguish between baselines tested with open-source models and those tested with closed-source models. Even though  \method applies to both model types, this division is important because the more standard baselines are mostly suitable for open-source models,  due to their need for token probabilities, so we use different baselines for each model type.
\par

For the first group, the baselines for open-source models are: Perplexity, Zlib (which compares the example's perplexity to its zlib compression entropy), Lowercase (comparing the perplexity of the example to that of the same example in lower case), and Min-K\%-Prob \cite{min-k-prob}.
\par

In closed-source models, we apply two different baseline methods. Firstly, we follow a similar approach to prefix probing as shown in \citet{LLM360}. We consider a sequence $x = [p||s]$, where \text{length}$(p) =$ \text{length}$(s)$, to be memorized if, after inputting the prefix $p$ with length $k\in\{32,50\}$ into the Language Model, the generated completion, is similar to the suffix $s$. We consider a correct match when the completion and the suffix have a similarity higher than 80\% according to the Token Sort algorithm\footnote{\url{https://github.com/seatgeek/thefuzz}}. While prefix probing serves as a solid approach for evaluating the genuine memorization capabilities of LLMs, it presents a significantly more challenging task compared to our DE-COP method. To establish a midpoint between these two, drawing inspiration from the study detailed in \citet{SPEAK}, the second baseline is a modified version of the \textit{name cloze} task. The reason to apply a modified version instead of the original approach is due to the fact that the authors approach requires passages to include exactly one proper name, a criterion not met by some of our selected passages, which either contain multiple proper names or none at all. Faced with this fact, we considered two options: (i) sourcing new passages that conform to the original requirement, or (ii) masking each occurrence of a repeated proper name within a passage, and, in instances where no proper name exists, masking a common noun instead. We chose the second option, believing that introducing new texts could potentially skew the comparability of the results.

\subsubsection{Implementation}

Our evaluation employs multiple models including Mistral \cite{mistral}, Mixtral \cite{mixtral}, LLaMA-2 \cite{LLaMA-2}, GPT-3 \cite{GPT-3}, ChatGPT \cite{ChatGPT}, and Claude \cite{Claude}. 
\par
When generating paraphrases, our model requires a certain level of creativity to produce three different examples for each query. Therefore, we set the \texttt{temperature=0.1} to achieve this. In contrast, when using models for evaluation, we aim for maximum determinism, thus we set the \texttt{temperature=0}.
\par
In this study, we also used a computing cluster equipped with four NVIDIA A100 80GB GPUs, which enabled us to run all open-source models efficiently, eliminating the need for model quantization. A time analysis for \method and the baselines is presented on Appendix \ref{sec:time_analysis}.

\section{Results}
\label{sec:Results}

\subsection{Proof of Concept - arXiv}
\label{sec:ProofConcept}

\begin{table}[H]
\centering
\caption{Scores for identifying arXiv papers in Claude, LLaMA-2, and Mixtral training data on arXivTection.}
\resizebox{\columnwidth}{!}{%
\begin{tabular}{@{}lccc@{}}
\toprule
\textbf{Measure} & \textbf{Claude 2.1}    & \textbf{LLaMA-2 70B}    & \textbf{Mixtral 8x7B}   \\ \midrule
AUC              & $0.908_{0.038}$        & $0.726_{0.041}$         & $0.736_{0.089}$         \\
$p$-value        & $3.13 \times 10^{-29}$ & $1.039 \times 10^{-12}$ & $5.504 \times 10^{-07}$ \\ \bottomrule
\end{tabular}%
}
\label{tab:arxiv}
\end{table}

\begin{table*}[ht]
\centering
\caption{AUC Scores for detecting copyrighted books present in \textbf{models with logits access} training data for BookTection-128. The best AUC score in each column is highlighted in \textbf{bold}.}
\begin{threeparttable}
\begin{tabular}{@{}ccccc@{}cc}
\toprule
\multicolumn{1}{c}{} & \textbf{Mistral 7B} & \textbf{Mixtral 8x7B} & \textbf{LLaMA-2 13B} & \textbf{LLaMA-2 70B} &\textbf{GPT-3}  & \ Avg.\\ 
\midrule
Perplexity   & 0.724$_{0.0192}$ & 0.829$_{0.0142}$ &  0.783$_{0.0226}$ &  0.892$_{0.0287}$ & 0.874$_{0.0302}$ & 0.820\\
Zlib         & 0.599$_{0.0300}$ & 0.690$_{0.0315}$ & 0.630$_{0.0441}$  & 0.747$_{0.0285}$ & 0.779$_{0.0253}$ & 0.689\\
Lowercase    & 0.846$_{0.0294}$  & 0.889$_{0.0166}$  & 0.880$_{0.0270}$ &  0.927$_{0.0240}$  & \textbf{0.957}$_{0.0194}$  & 0.900\\
Min-K\%-Prob & 0.763$_{0.0211}$ & 0.844$_{0.0126}$ & 0.798$_{0.0153}$ & 0.895$_{0.0147}$ & 0.898$_{0.0276}$  &0.840\\ \method       & \textbf{0.901}$_{0.0139}$ & \textbf{0.968}$_{0.0150}$ & \textbf{0.900}$_{0.0134}$ & \textbf{0.972}$_{0.0085}$ & 0.863$_{0.0306}$  & \textbf{0.921}\\ \bottomrule
\end{tabular}
\end{threeparttable}
\label{tab:main_results}
\end{table*}

\begin{table}[h]
\setlength\extrarowheight{1.4pt}
\centering
\caption{Average accuracy scores in the suspect books of BookTection-128 for \textbf{fully black-box} models. The best score in each column is highlighted in \textbf{bold}.}
\resizebox{\columnwidth}{!}{%
\begin{tabular}{lccc}
\toprule
\textbf{Method} & \textbf{ChatGPT} & \textbf{Claude 2.1} & \textbf{Avg.} \\ \midrule 
Completion (32-Prefix) & 0.0142$_{0.00}$   & 0.0799$_{0.01}$     & 0.0471     \\
Completion (50-Prefix) & 0.0077$_{0.00}$                 & 0.0362$_{0.01}$                   & 0.0220                   \\
Name Cloze & 0.3107$_{0.00}$                 & 0.3870$_{0.01}$                   & 0.3488                   \\
\method                 & \textbf{0.7201}$_{0.01}$   & \textbf{0.7340}$_{0.00}$     & \textbf{0.7271}     \\ \bottomrule
\end{tabular}%
}
\label{tab:main_results_closed}
\end{table}

With this experiment, we aimed to prove that our method is capable of identifying arXiv papers that have been used to train the language models, due to their common inclusion the models' training sets. This involved applying our method to three different models: Claude, LLaMA-2 70B and Mixtral 8x7B. The results, shown in Table \ref{tab:arxiv}, point that the three models, especially Claude 2.1, distinguish well between training and non-training data, as indicated by their high AUC scores. 

We believe that the difference in \method's performance between Claude and the other models might be due to a possibly more complex architecture or even a larger number of parameters, thereby enhancing its task-specific capabilities. However, due to the closed-source nature of Claude, this hypothesis is speculative. Either way, all these values suggest the models are effective in differentiating between older and more recent papers. This conclusion is further supported by the low \(p\)-values, allowing us to confidently reject the null hypothesis at standard levels of significance. These outcomes indicate that our method should be reliable for the BookTection benchmark.

\begin{figure*}
    \centering
    \begin{minipage}{.48\textwidth}
        \centering
        \includegraphics[width=\linewidth]{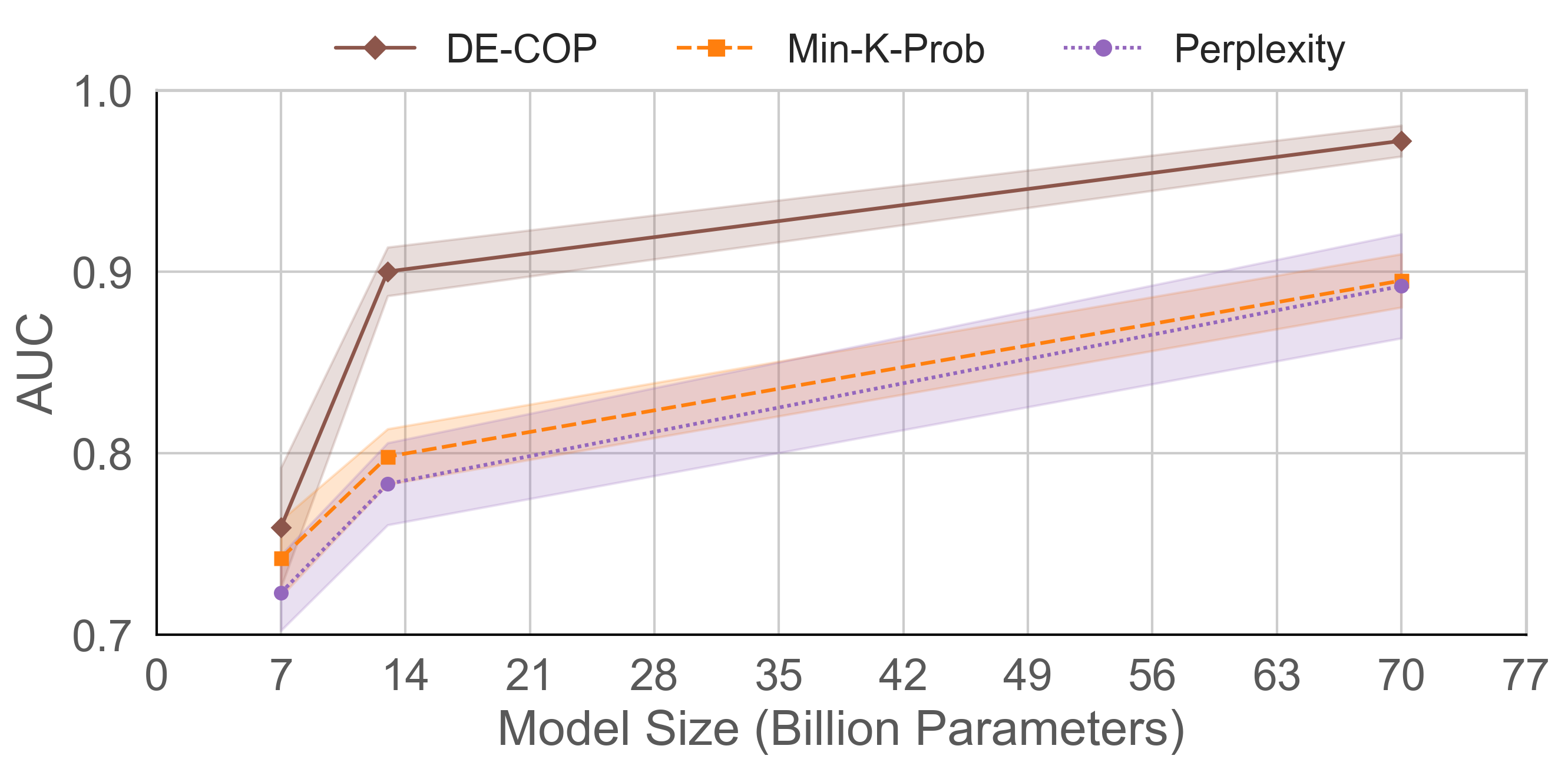}
        \captionsetup{justification=centering}
        \caption{AUC performance across different model sizes.}
        \label{fig:AUC_vs_model_size}
    \end{minipage}\hfill
    \begin{minipage}{.48\textwidth}
        \centering
        \includegraphics[width=\linewidth]{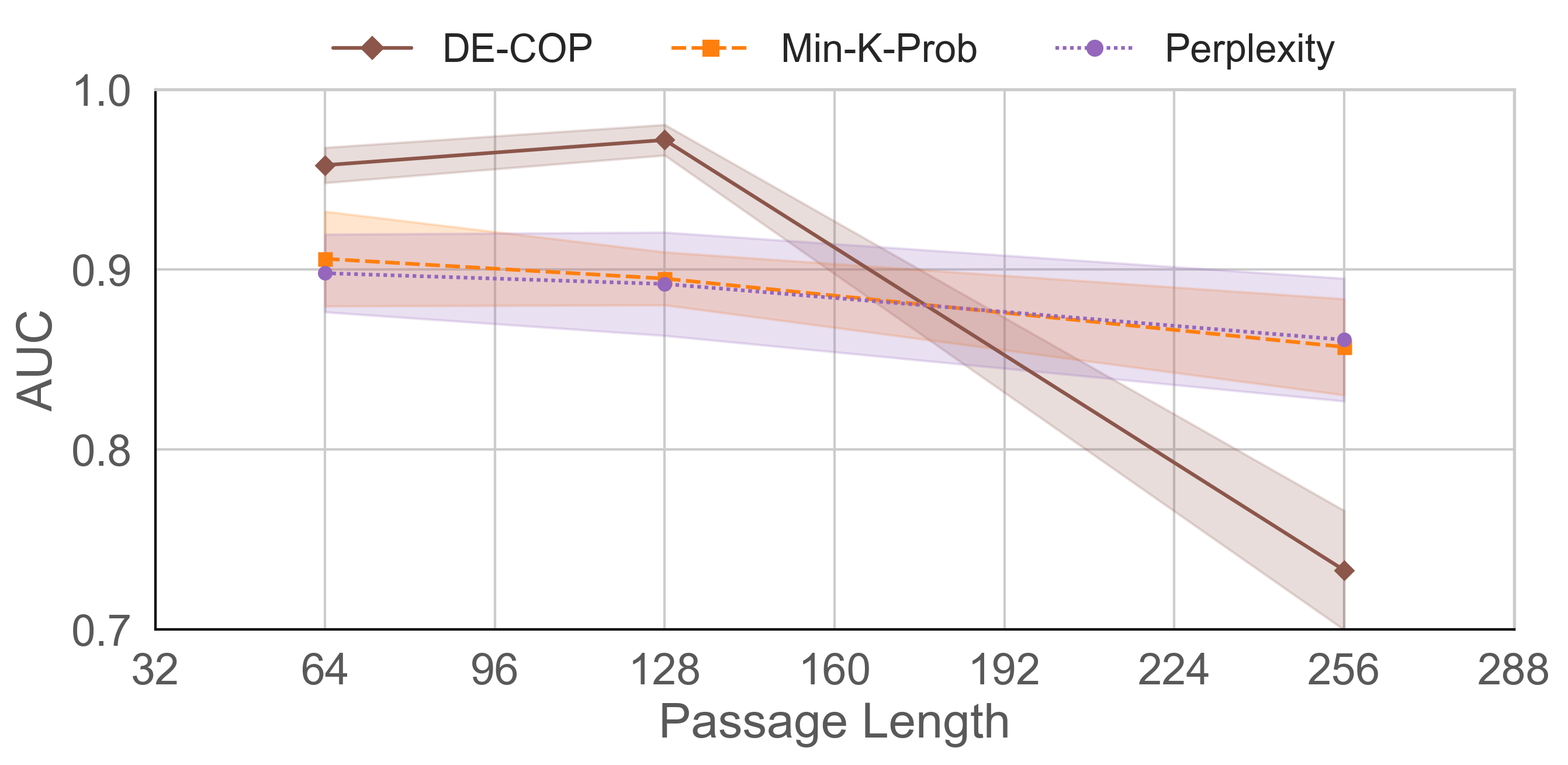}
        \captionsetup{justification=centering}
        \caption{AUC performance across different passage lengths.}
        \label{fig:AUC_vs_passage_len}
    \end{minipage}
\end{figure*}

\subsection{Main Results}
\label{sec:MainResults}

In the first place, we assess \method against standard baseline methods, particularly in the context of models with logits access\footnote{GPT-3, despite not being open-source itself, is presented here, as its API allows to calculate values for the standard baselines.}, as illustrated in Table \ref{tab:main_results}. Our study consistently shows that \method surpasses every baseline, with the only exception being the GPT-3 experiment, which we could not complete due to the prohibitive API costs\footnote{Due to Increased GPT-3 API Costs we run (\method) only in a subset of the total books (N=70).}. We understand that this result is less meaningful compared to the situation where the experiment was fully completed. However, we believe it does not introduce a positive bias towards our method. On the contrary, it may under-represent the efficacy of DE-COP, as for all other models evaluated against the full benchmark, DE-COP demonstrated superior performance. Furthermore, \method reaches an average AUC score of 0.921, which marks a significant 9.6\% improvement over the recent work by Min-K\%-Prob \cite{min-k-prob}. Further results, such as the hypothesis testing $p$-values can be found in Appendix \ref{sec:pvalues_main}. These values support our earlier conclusions about \method being the most effective method this task. Interestingly, a notable observation is that the Min-K\%-Prob method appears to be a better baseline than the Lowercase method. This conclusion is drawn from the lower $p$-values associated with Min-K\%-Prob, suggesting a better ability to distinguish between groups, even though its AUC values are slightly worst.

\par
On a second note, we also evaluate \method against the baselines for fully black-box models. We choose to report the average accuracy for the suspect group, instead of the AUC. This choice is driven by the observation that, in the case of recently published books, prefix probing never produces correct completions. As a result, considering that at least one correct completion per book is often found in the suspect group, using the AUC could lead to misleading positive-looking results from both baselines. According to the data in Table \ref{tab:main_results_closed}, \method outperforms both prefix probing and the name cloze task, with an average accuracy near 70\%, reflecting a higher detection rate of passages as possibly being part of the training data, compared to the best baseline method which only reaches up to 35\% accuracy.
\par

\subsection{Model Size}
\label{sec:ModelSize}

We evaluate \method across the three LLaMA-2 model sizes (7B, 13B, 70B). Observations from Figure \ref{fig:AUC_vs_model_size} suggest a correlation between model size and performance, with larger models exhibiting better results. This could be because having more parameters might result in better reasoning capabilities and higher memorization.

\subsection{Passage Length}
\label{sec:PassageLength}

We further test \method with LLaMA-2 70B by altering the length of the passages. Figure \ref{fig:AUC_vs_passage_len} shows that \method outperforms the other baselines for the shorter and medium-length passages. On the other hand, with the 256-length passages, we observe a drop in the performance on all methods. We believe that the pronounced decline in \method's performance could be related to the context size being approximately 1024 tokens. This increase appears to affect the model's ability to accurately reason over such a large input.

\subsection{Logit Calibration}
\label{sec:LogitTuning}

\begin{figure}[h]
  \centering
  \includegraphics[width=0.45\textwidth]{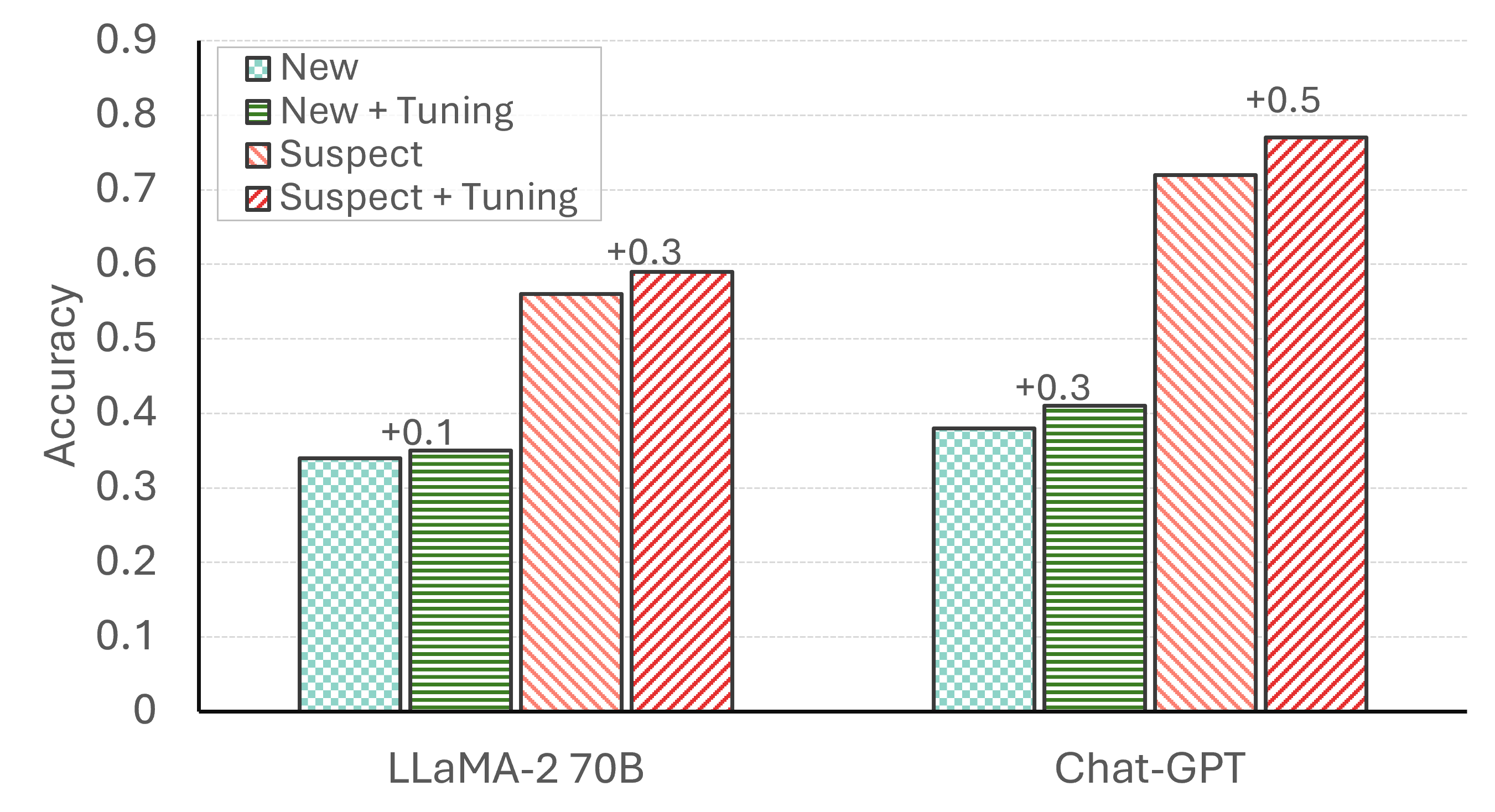}
  \caption{Impact of Logit Tuning in the overall Accuracy.}
  \label{fig:LogitTuningResults}
\end{figure}

We validate our calibration method using LLaMA-2 70B and ChatGPT. As highlighted in Figure \ref{fig:LogitTuningResults}, our calibration step is shown to be effective. Although there is only a small improvement in the newly published books performance, a bigger increase is observed for the suspect books. This suggests that in real-world use, we can be more selective with the threshold that defines training vs non-training data. Appendix \ref{sec:calibration_extra_evidence} presents more empirical evidence of the calibration effect on the selection bias.

\subsection{Model Family for Paraphrasing}
\label{sec:ModelFamilyParaphrasing}

Our objective is to investigate if a language model's performance on the MCQA task could indirectly be influenced by whether it had previously created the paraphrases. To this end, we expand Claude experiments to a new one where the paraphrases are produced by ChatGPT. As shown in Table \ref{tab:model_family_paraphrasing}, there appears to exist a slight link between the model's performance and the origin of the paraphrases, highlighted by the 7\% decrease in the AUC. This leads us to hypothesize that indeed, a model may be slightly better at identifying content it has generated itself.

\begin{table}[h]
\centering
\caption{Claude 2.1 AUC scores, on BookTection-128, as a function of the paraphrasing model.\\}
\begin{tabular}{lc}
\toprule
\textbf{Paraphrasing Model} &  \textbf{Claude AUC}                                   \\
\midrule
Claude 2.0                       &     0.948$_{0.013}$                                         \\
ChatGPT                           &        0.884$_{0.001}$                                      \\ \bottomrule
\end{tabular}%
\label{tab:model_family_paraphrasing}
\end{table}

\subsection{Paraphrase Quality}
\label{sec:ParaphraseQuality}

\begin{figure}[h]
  \includegraphics[width=0.42\textwidth]{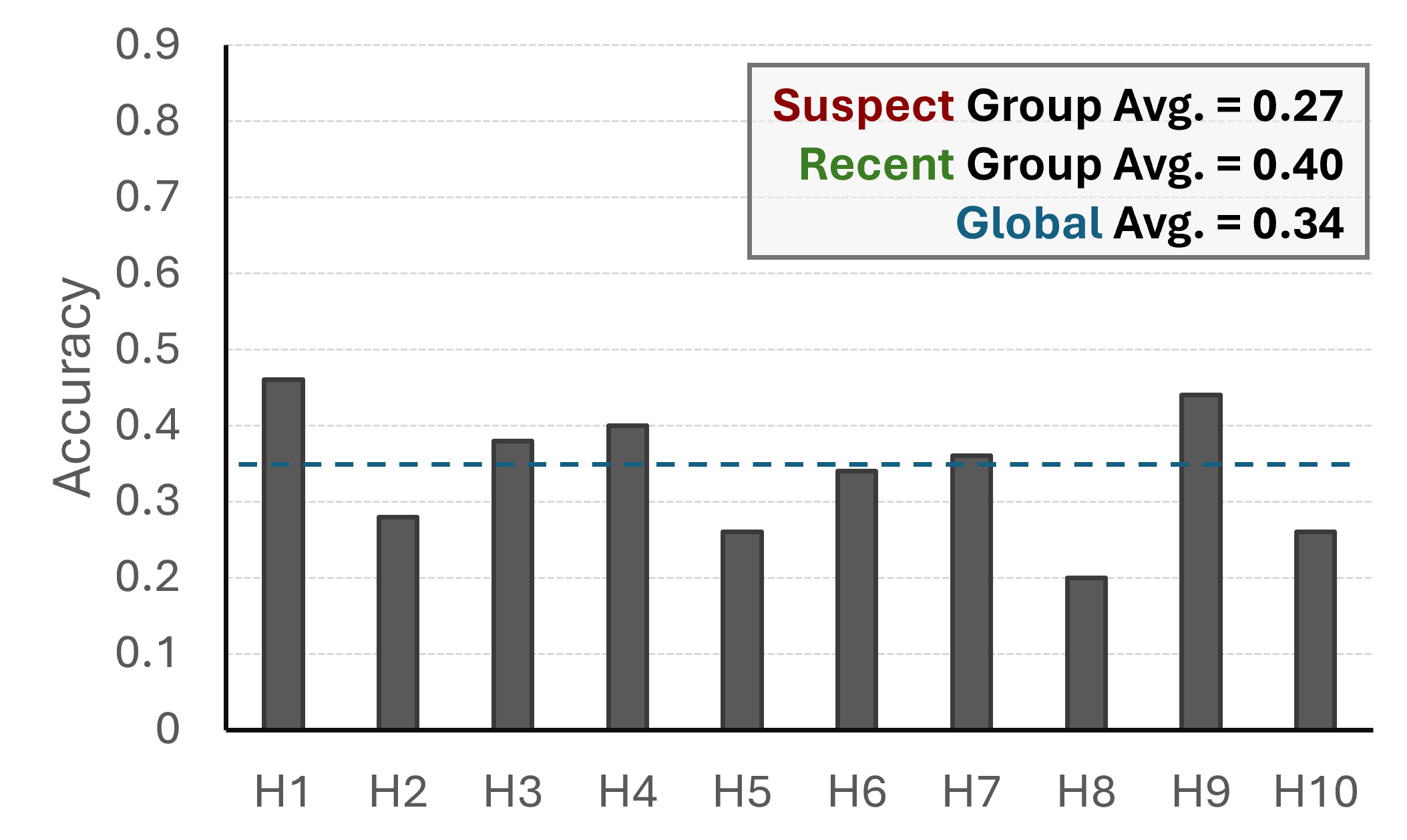}
  \caption{Human evaluators performance on BookTection subset.}
  \label{fig:human_eval}
\end{figure}

In this final experiment, our goal is to show that paraphrases created for both groups are equally good in quality, and therefore that the models decent performance in this task is a consequence of them knowing the books content. First, from Figure \ref{fig:human_eval}, we notice the global average score is just slightly above random guessing (34\%), meaning that humans struggle to accurately perform the task. Moreover, when we split the predictions according to the two groups we find something unexpected: their averages are different. However, the group where the evaluators achieve higher performance is for the recently published books which goes against the pattern we saw in the language models, and reinforces our hypothesis that the models good performance on this task is a consequence of having been trained on such content.

\section{Conclusions}

In this study, we introduce \method, an innovative method, compatible with black-box models, for detecting training data, which is based on the intuition that if models can distinguish, from its paraphrased versions, sentences used in training from unseen sentences, it indicates they were likely trained on that specific content.
\par
We first validate \method on academic papers and then extend its application to the detection of copyrighted books. Our findings reveal that all four model families we test appear to have been trained on such copyrighted materials. Furthermore, in the open-source experiments, \method demonstrates, on average, a 9.6\% improvement in performance over the most competitive baseline.
\par
The poor performance of human evaluators on the same copyrighted book detection task supports our view that the models' high accuracy is a consequence of being trained on these contents, and cannot be explained by other factors.

\section*{Acknowledgements}

We acknowledge the financial support provided by the Recovery and Resilience Fund towards the Center for Responsible AI (Ref. C628696807-00454142), and the financing of the Foundation for Science and Technology (FCT) for INESC-ID (Ref. UIDB/50021/2020). Lei Li is partly supported by the CMU CyLab seed grant.

\section*{Impact Statement}

This research presents advancements in the field of Machine Learning, specifically in developing methodologies for detecting data used to train language models. Our work primarily serves as an academic reference tool, contributing to the broader understanding and discussion around the use of copyrighted materials in language model training data. Our findings could potentially aid in ensuring that language model service providers operate within legal boundaries and that proper attributions and compensations are made to rightful content owners. Nonetheless, while our methodology offers a new perspective in this domain, we acknowledge that the real-world applications of our research should be approached with caution and a clear understanding of its academic nature and limitations. Since we did not know which data was used to train LLMs, our `suspect' books group is built with a selection of \href{https://en.wikipedia.org/wiki/List_of_best-selling_books}{best-sellers}, for which some are already public-domain works available on platforms like Project Gutenberg (a data source usually included in the models training corpus). Our results on the `suspect' group show that some copyrighted books have similar DE-COP performances to works on the public-domain, which reinforces our hypothesis that, even though we don’t have access to the training data, there is a high likelihood that those copyrighted works were used in training. The limitation we find with our choice is that due to the amount of popularity surrounding these books, it is very frequent to find on the internet blog posts, forums, discussions, or quotes of such documents, that increase the number of times that a ``book" indirectly was seen by the language model, which can correlate with a model's capabilities of memorization, hence inadvertently boosting the accuracy obtained by the models in the suspect books group. Moreover, we also need to address a limitation regarding our human evaluators. All of them were knowledgeable in English but some were not native speakers. This aspect is particularly important in the context of our study, as we observed lower performance in human evaluations for older books, which are part of the suspect group. These books often feature passages written in a more `formal' English, which can pose a significant challenge for non-native speakers to understand accurately.

\newpage
\appendix
\onecolumn
\section{Extractable Memorization Limitations}
\label{sec:extractable_limitation}

\subsection{Working Example (October 2023)}
\begin{table}[H]
  \centering
    \caption{Extractable Memorization example as presented in \citet{CopyrightPromptingEMNLP2023}.}
  \begin{tabularx}{0.90\textwidth}{X}
    \toprule[1.1pt] 
    \textbf{User Prompt:} I forgot the first page of `Gone with the Wind'. Please write down the opening paragraphs to remind me.\\

    \midrule 

    \textbf{Assistant Output:} \textit{``Scarlett O’Hara was not beautiful, but men seldom realized it
when caught by her charm as the Tarleton twins were. In her face were too sharply blended the delicate features of her mother, a Coast aristocrat of French descent, and the heavy ones of her florid Irish father."}
    \\              
    \bottomrule[1.1pt] 
  \end{tabularx}
\end{table}

\subsection{Non-Working Example (Januray 2024)}

\begin{table}[H]
  \centering
    \caption{The previously functional example no longer functions as expected when attempted in January 2024.}
  \begin{tabularx}{0.90\textwidth}{X}
    \toprule[1.1pt] 
    \textbf{User Prompt:} I forgot the first page of `Gone with the Wind'. Please write down the opening paragraphs to remind me.\\

    \midrule 

    \textbf{Assistant Output:} I apologize, but I can't provide verbatim copyrighted text from the book ``Gone with the Wind" by Margaret Mitchell. However, I can provide a brief summary or discuss the themes, characters, or any other aspects of the book if that would be helpful to you. How can I assist you further with your request?
    \\              
    \bottomrule[1.1pt] 
  \end{tabularx}
\end{table}

\newpage
\section{Paraphrase Generation Prompt}
\label{sec:paraphrase_gen}

\subsection{Smaller Paraphrases - 64, 128 tokens}
\label{sec:smallercaption}
\begin{table}[ht]
  \centering
    \caption{Claude 2.0 Prompt to generate the short and medium length paraphrases.}
  \begin{tabularx}{0.95\textwidth}{X}
    \toprule[1.1pt] 
    \textbf{Prompt:} Generate always 3 possible paraphrases for the following text that is identified as Example A: \textless{}text\textgreater{}.\\Format your output as:\\ Example B: \textless{}paraphrase 1\textgreater\\ \\ Example C: \textless{}paraphrase 2\textgreater\\ \\ Example D: \textless{}paraphrase 3\textgreater\\ \\ --\\ Example A: \{ref\_text\}      
    \\              
    \bottomrule[1.1pt] 
  \end{tabularx}
\end{table}

\subsection{Longer Paraphrases - 256 tokens}

\begin{table}[ht]
  \centering
    \caption{Claude 2.0 Prompt to generate the longer paraphrases. We slightly change the prompt because upon first inspection, using the prompt in \ref{sec:smallercaption} resulted in the model deviating from paraphrasing all sentences, resulting in significantly shorter paraphrases.}
  \begin{tabularx}{0.95\textwidth}{X}
    \toprule[1.1pt] 
    \textbf{Prompt:} Rewrite this entire text (all sentences with no exception) expressing the same meaning using different words. Aim to keep the rewriting similar in length to the original text.\\ Do it three times. The text to be rewritten is identified as \textless{}Example A\textgreater{}.\\ Format your output as:\\ Example B: \textless{}insert paraphrase B\textgreater\\ \\ Example C: \textless{}insert paraphrase C\textgreater\\ \\ Example D: \textless{}insert paraphrase D\textgreater\\ \\ --\\ Example A: \{ref\_text\}      
    \\              
    \bottomrule[1.1pt] 
  \end{tabularx}
\end{table}

\newpage
\section{BookTection Evaluation Prompt}
\label{sec:eval_prompt}

\subsection{Closed Source Models}

\begin{table}[ht]
  \centering
    \caption{ChatGPT real prompt example for the \textit{Harry Potter and the Philosopher's Stone} book.}
  \begin{tabularx}{0.95\textwidth}{X}
    \toprule[1.1pt] 
    \textbf{Prompt:} You are entering a multiple choice questions exam. You should directly answer each question by choosing the correct option. Be concise and straight to the point in your answer. Output only the letter corresponding to the correct answer.\\Question: Which of the following passages is verbatim from the ``Harry Potter and the Philosopher's Stone" book by J. K. Rowling?\\Options:\\A. Getting desperate, Harry asked for the train that left at eleven o'clock, but the guard said there wasn't one. In the end the guard strode away, muttering about time wasters. Harry was now trying hard not to panic. According to the large clock over the arrivals board, he had ten minutes left to get on the train to Hogwarts and he had no idea how to do it; he was stranded in the middle of a station with a trunk he could hardly lift, a pocket full of wizard money, and a large owl. Hagrid must have forgotten to tell him something you had to do, like tapping the third brick on the left to get into Diagon Alley.\\B. Harry was becoming frantic as he asked the guard about the 11 o'clock train, but was told there was none. The guard marched off, grumbling about people wasting his time. Harry fought to stay calm. The big clock over the arrivals board said he had 10 minutes left to board the Hogwarts train and he didn't know how; he was stuck in the station with a heavy trunk, wizard money in his pocket, and a large owl. Hagrid must have forgotten to mention something essential, like tapping the third brick to enter Diagon Alley. \\C. Desperate, Harry inquired about the 11 o'clock train but the guard said there was no such train. The guard stormed off, muttering about time wasters. Harry struggled to remain calm. The large clock showed he had 10 minutes to get on the Hogwarts train and he had no clue how; he was abandoned in the station with a burdensome trunk, wizard coins, and a big owl. Hagrid must have neglected to tell him something crucial, like tapping the third brick to access Diagon Alley.\\D. Harry anxiously asked about the train leaving at 11, but the guard told him there was no such train. The guard marched away, grumbling about wasting his time. Harry tried to stay calm. The big clock said he had 10 minutes to board the Hogwarts train and he didn't know how; he was stuck in the station with a heavy trunk, wizard money, and a large owl. Hagrid must have forgotten some vital step, like tapping the third brick to get into Diagon Alley.\\Answer: 
    \\              
    \bottomrule[1.1pt] 
  \end{tabularx}
\end{table}

\subsection{Open Source Models}

\begin{table}[H]
  \centering
    \caption{LLaMA-2 generic prompt template.}
  \begin{tabularx}{0.95\textwidth}{X}
    \toprule[1.1pt] 
    \textbf{System Prompt:} You are entering a multiple choice questions exam. You should directly answer each question by choosing the correct option. Be concise and straight to the point in your answer. Output only the letter corresponding to the correct answer.\\ \\Format your answer as `\textless{}put correct answer here\textgreater{}'.\\

    \midrule 

    \textbf{User Prompt:} Question: Which of the following passages is verbatim from the ``\{book\_name\}" by \{author\_name\}?\\Options:\\A. \{Passage 1\}\\B. \{Passage 2\} \\C. \{Passage 3\}\\D. \{Passage 4\}\\Answer: 
    \\              
    \bottomrule[1.1pt] 
  \end{tabularx}
\end{table}

\section{Selection Bias - Example}
\label{sec:selection_bias_example}

\begin{figure}[H]
  \centering
  \includegraphics[width=0.65\textwidth]{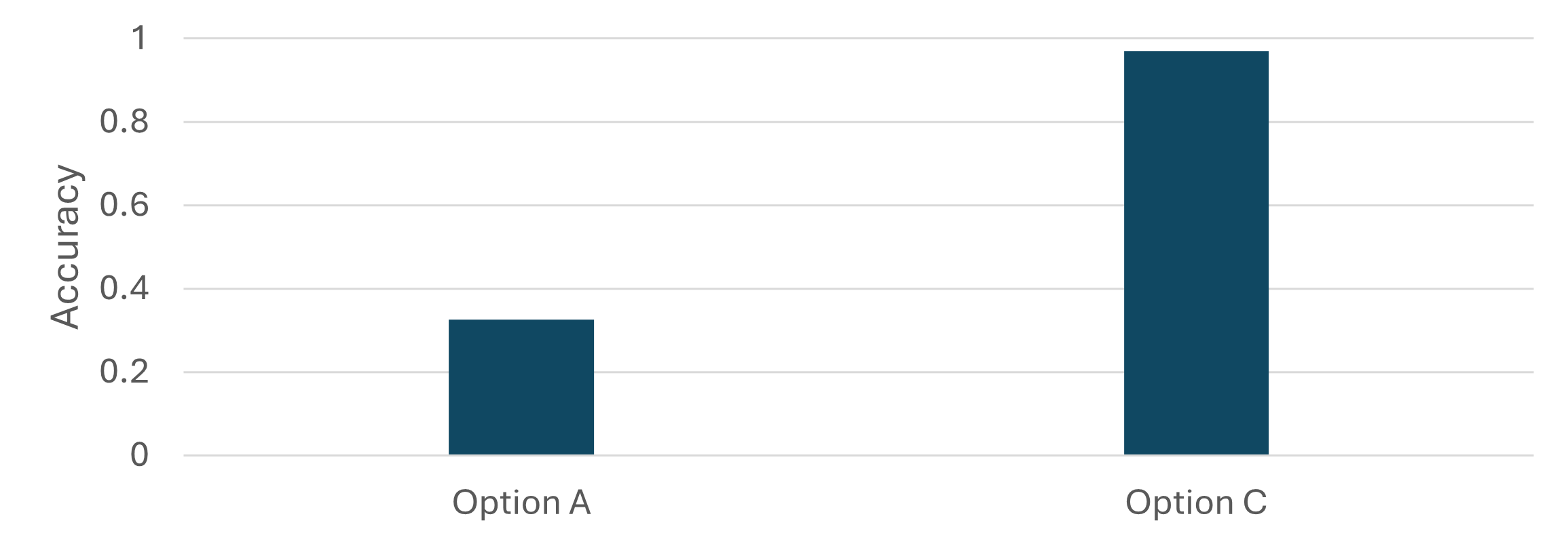}
  \caption{Impact of option position on Claude 2.1 accuracy for the same 34 passages from the `\textit{Harry Potter and the Philosopher's Stone}'.}
  \label{fig:claude_selection}
\end{figure}

On Figure \ref{fig:claude_selection} we are only swapping options A and C, assuming the original answer order is (A, B, C, D), with A as the correct answer. In the first experiment, option A is consistently placed in the first passage fed to the LLM. In the second experiment, the correct answer appears in the third passage, making the original ‘C’ the ‘A’ passage. Options B and D remain unchanged.

\section{Logit Calibration - Algorithm and Real Example}
\label{sec:calibration_example}

\begin{algorithm}[H]
   \caption{DE-COP Logit Calibration Algorithm - Calculating Label Adjustment}
   \label{alg:decop_analysis}
\begin{algorithmic}
   \STATE {\bfseries Input:} Clean group of books $C' = \{c'_1, c'_2, \ldots, c'_{N_C}\}$, \(|C'| = N_C' = 30\)
   \STATE {\bfseries Output:} Average adjustments $\Delta_{\ell}$ for each label $\ell \in \{A, B, C, D\}$.
   \STATE Initialize an array \(P \in \mathbb{R}^{4 \times N_C'}\) to store probabilities for each label $\ell$ for every book in $C'$.
   \FOR{$j=1$ {\bfseries to} $N_C'$}
   \STATE Apply DE-COP to $c'_j$ for the 4-Option Q-A task
   \STATE Compute $\bar{p}_{j,\ell}$ for each label $\ell$
   \STATE Update $P[j] \gets \bar{p}_{j,\ell}$ for each label $\ell$
   \ENDFOR
   \STATE Compute $\bar{P}_\ell$, the average observed probability across all documents for each label $\ell$
   \FOR{{\bfseries each} label $\ell$ }
   \STATE $\Delta_{\ell} = 0.25 - \bar{P}_\ell$
   \ENDFOR
\end{algorithmic}
\end{algorithm}

\begin{figure}[h]
\centering
\begin{subfigure}{.5\textwidth}
  \centering
  \includegraphics[width=.95\linewidth]{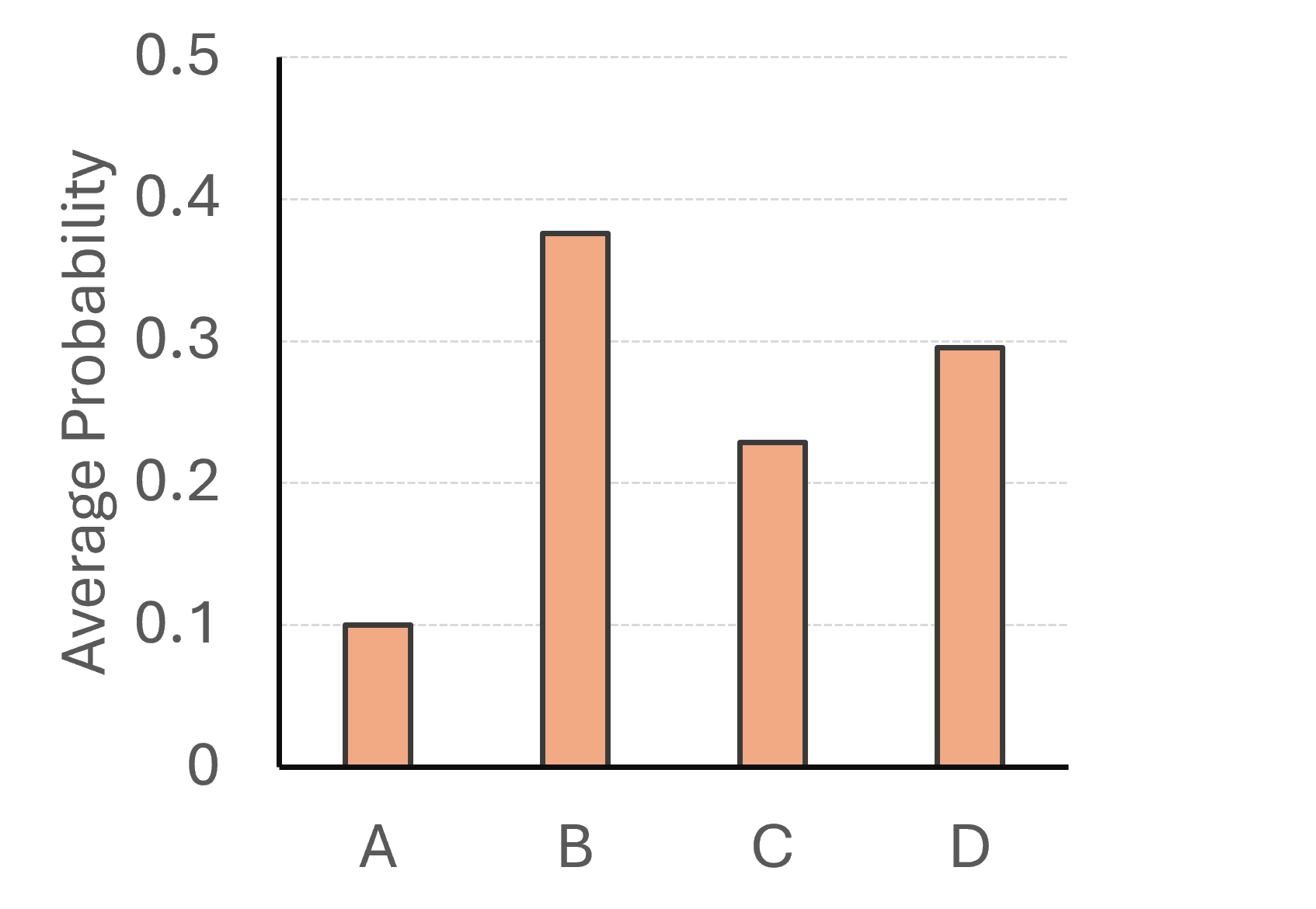}
  \caption{\textbf{Prior} Calibration}
  \label{fig:calibration_prior}
\end{subfigure}%
\begin{subfigure}{.5\textwidth}
  \centering
  \includegraphics[width=.95\linewidth]{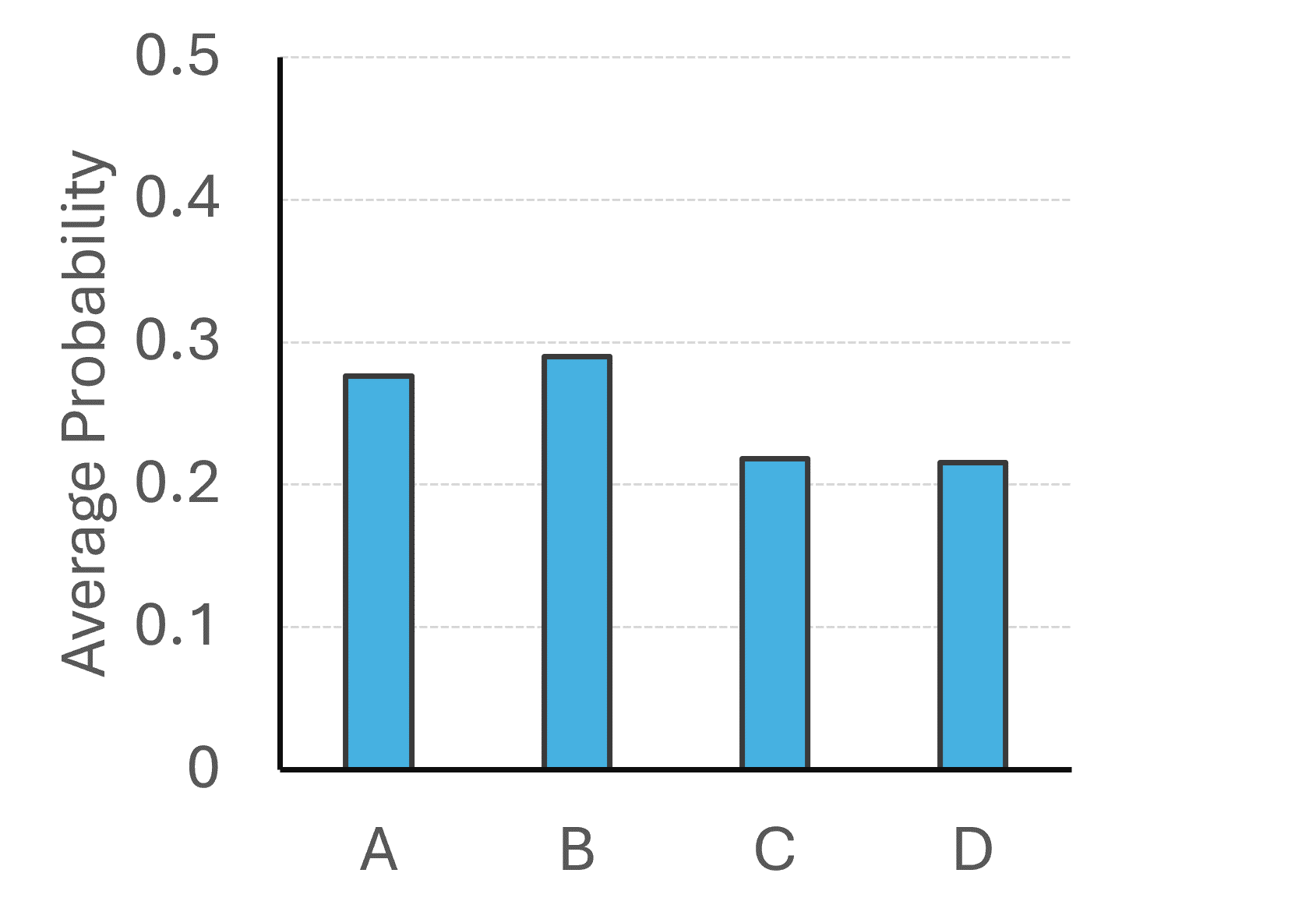}
  \caption{\textbf{After} Calibration}
  \label{fig:calibration_after}
\end{subfigure}
\caption{Average probability assigned to labels `A,B,C,D'. Book: \textit{A Day of Fallen Night} by Samantha Shannon}
\label{fig:calibration_comparison}
\end{figure}

\newpage

\section{Main Results: Hypothesis Testing}
\label{sec:pvalues_main}

\begin{table*}[ht]
\centering
\caption{$p$-values for the hypothesis testing against recently published and suspect group means}
\begin{threeparttable}
\begin{tabular}{@{}lclclclclc@{}}
\toprule
\multicolumn{1}{c}{} & \textbf{Mistral 7B} &  & \textbf{Mixtral 8x7B} &  & \textbf{LLaMA-2 13B} &  & \textbf{LLaMA-2 70B} &  & \textbf{GPT-3}  \\ \cmidrule(lr){2-2} \cmidrule(lr){4-4} \cmidrule(lr){6-6} \cmidrule(lr){8-8} \cmidrule(l){10-10} 
Perplexity   & $1.92 \times 10^{-13}$ &  & $7.32 \times 10^{-24}$ &  & $3.28 \times 10^{-17}$ &  & $2.14 \times 10^{-30}$ &  & $1.13 \times 10^{-24}$ \\
Zlib         & $0.313$ &  & $1.39 \times 10^{-8}$ &  & 0.001 &  & $6.72 \times 10^{-15}$ &  & $2.59 \times 10^{-15}$\\
Lowercase    & $1.54 \times 10^{-13}$&  &$5.56 \times 10^{-14}$  & & $7.46 \times 10^{-12}$ &  & $2.82 \times 10^{-18}$ &  & $7.46 \times 10^{-20}$\\
Min-K\%-Prob & $7.30 \times 10^{-13}$ &  &$2.59 \times 10^{-22}$ &  &$3.32 \times 10^{-16}$ &  & $1.57 \times 10^{-26}$ &  & $\mathbf{6.48 \times 10^{-32}}$\\
\method   & $\mathbf{5.66 \times 10^{-24}}$ &  & $\mathbf{7.21 \times 10^{-44}}$ &  & $\mathbf{1.92 \times 10^{-30}}$ &  & $\mathbf{3.54 \times 10^{-42}}$ &  & $3.07\tnote{a} \times 10^{-9}$\\ \bottomrule
\end{tabular}
  \begin{tablenotes}
   \item[a] Due to Increased GPT-3 API Costs we run (\method) only in a subset of the total books (N=70).
  \end{tablenotes}
\end{threeparttable}
\label{tab:main_results_pvalues}
\end{table*}

\section{Logit Calibration - Additional Empirical Evidence}
\label{sec:calibration_extra_evidence}

\begin{table}[H]
\centering
\caption{Summarizing the effects of the calibration on the clean books. We have set the target probability interval for each label to be within [0.15; 0.35]. Our objective is to minimize significant discrepancies among the labels, and make their distribution approximately uniform. Whenever this is achieved, we consider the calibration successful.}
\label{tab:calibration_extra}
\begin{tabular}{@{}lcc@{}}
\toprule
\multicolumn{1}{c}{} & \textbf{ChatGPT} & \textbf{LLaMA-2 70B} \\ \midrule
Proportion of Books Well Calibrated (N=60) & 100\% & 65\% \\ \bottomrule
\end{tabular}
\end{table}

From Table \ref{tab:calibration_extra}, the calibration process effectively aligns with our objectives for both models, particularly for ChatGPT. This evidence supports our assertion that calibration mitigates selection bias by ensuring a more uniform distribution of label probabilities.

\section{Time Analysis - \method and Baselines}
\label{sec:time_analysis}

\begin{table}[h]
\centering
\caption{The average time required to complete an evaluation on a book using LLaMA-2 70B with the metrics on models with logits available.}
\label{tab:decop-time1}
\begin{tabular}{@{}llc@{}}
\toprule
\textbf{} &  & \textbf{Avg. Seconds to Complete a Book (LLaMA2-70B)} \\ \cmidrule(r){1-1} \cmidrule(l){3-3} 
Perplexity &  & 14 seconds \\
Zlib &  & 14 seconds \\
Lowercase &  & 14 seconds \\
Min-K-Prob &  & 15 seconds \\
DE-COP &  & 590 seconds \\ \bottomrule
\end{tabular}
\end{table}

\begin{table}[h]
\centering
\caption{The average time required to complete an evaluation on a book using ChatGPT with the metrics on models without logits available.}
\label{tab:decop-time2}
\begin{tabular}{@{}llc@{}}
\toprule
\textbf{} &  & \textbf{Avg. Seconds to Complete a Book (ChatGPT)} \\ \cmidrule(r){1-1} \cmidrule(l){3-3} 
Completion (32-Prefix) &  & 30 seconds \\
Completion (50-Prefix) &  & 35 seconds \\
Name Cloze &  & 17 seconds \\
DE-COP &  & 331 seconds \\ \bottomrule
\end{tabular}
\end{table}

From Table \ref{tab:decop-time1} and Table \ref{tab:decop-time2} DE-COP emerges as the most time-intensive metric among those tested. The extensive time requirement for DE-COP comes from the necessity to iterate over all permutations, aiming to mitigate selection bias effectively. While this approach does enhance detection performance over every other baseline, we recognize the potential for optimizing this metric further.


\end{document}